\documentclass[runningheads]{llncs}
\usepackage{graphicx}
\usepackage{float}
\usepackage[ruled,vlined]{algorithm2e}
\usepackage{multirow}

\begin{document}
\title{Adversarial Attacks on Deep Learning Systems for User Identification based on Motion Sensors}
%
%
\author{Cezara Benegui
		\and
        Radu Tudor Ionescu\vspace{-0.2cm}}

\institute{
University of Bucharest, 14 Academiei, Bucharest, Romania\\
\email{cezara.benegui@fmi.unibuc.ro, raducu.ionescu@gmail.com}
}

\titlerunning{Adversarial Attacks on Deep Learning Systems for User Identification}
\maketitle              
\begin{abstract}
For the time being, mobile devices employ implicit authentication mechanisms, namely, unlock patterns, PINs or biometric-based systems such as fingerprint or face recognition. While these systems are prone to well-known attacks, the introduction of an explicit and unobtrusive authentication layer can greatly enhance security. In this study, we focus on deep learning methods for explicit authentication based on motion sensor signals.
In this scenario, attackers could craft adversarial examples with the aim of gaining unauthorized access and even restraining a legitimate user to access his mobile device. To our knowledge, this is the first study that aims at quantifying the impact of adversarial attacks on machine learning models used for user identification based on motion sensors. To accomplish our goal, we study multiple methods for generating adversarial examples. We propose three research questions regarding the impact and the universality of adversarial examples, conducting relevant experiments in order to answer our research questions. Our empirical results demonstrate that certain adversarial example generation methods are specific to the attacked classification model, while others tend to be generic. We thus conclude that deep neural networks trained for user identification tasks based on motion sensors are subject to a high percentage of misclassification when given adversarial input.
\end{abstract}

%
%
\section{Introduction}

Nowadays, usage of mobile devices has grown exponentially, becoming the first choice for most users. Since personal devices allow manipulation and access to private data, security is one of the most important factors to consider. Whereas all operating systems and applications allow for standard security configuration, such as unlock patterns, PINs or facial recognition, it is well known that they are prone to attacks such as smudge~\cite{Aviv-WOOT-2010}, reflection~\cite{Xu-CCS-2013,Zhang-SPSM-2012} and video capture attacks~\cite{Shukla-CCS-2014,Simon-SPSM-2013,Ye-NDSS-2017}. With the advancement of technology, different sensors were introduced in  mobile devices, among others being motion sensors. Recent works \cite{Benegui-Access-2020,Buriro-SPW-2016,Buriro-ISBA-2017,Neverova-Access-2016,Sitova-TIFS-2016,Sun-ECML-2017} proposed continuous and unobtrusive authentication systems that use motion sensors, such as the gyroscope and the accelerometer, to silently identify users, employing machine learning models, e.g.~convolutional neural networks. Unobtrusive authentication methods, based on motion sensor data, are best used for enhancing implicit authentication systems (such as facial detection algorithms and fingerprint scanners). Due to recent scientific progress, implicit methods require fewer steps for the registration phase. As shown in literature \cite{Benegui-Access-2020}, machine learning models based on motion sensor data are great candidates for the aforementioned process, taking into consideration their capability to register users in as less as 20 steps. 

In this study, our objectives are $(i)$ to determine if adversarial examples can be crafted for deep learning models based on motion sensor signals, and $(ii)$ to evaluate the impact and the universality of adversarial examples for deep learning models based on motion sensor signals. 
In an adversarial setting, data samples are crafted by an attacker and fed as input to a classifier, with the aim of inducing a different output, other than the expected label. Although deep neural networks used in image recognition and computer vision tasks are known to be prone to adversarial attacks \cite{chen2017show,gragnaniello2018analysis,li2018robust}, to our knowledge, there is no research conducted on deep neural networks trained on motion signals collected from mobile devices. Therefore, we are the first to experiment with various adversarial attack generation methods on different neural network architectures, that are trained to identify users based on motion sensor signals. 
Within this frame of reference, we seek to gather empirical evidence on how different neural networks architectures are affected by the generated adversarial examples.
Our approach is to employ convolutional neural models trained for user identification \cite{Benegui-Access-2020}, using samples from HMOG data set \cite{Sitova-TIFS-2016}. HMOG contains samples recorded during screen tap gestures, from 50 users, while sitting. Samples include values collected from the accelerometer and gyroscope, each providing data points on 3-axis, namely $(x,y,z)$. We convert each signal into a gray-scale image by repeating the six one-dimensional signals, using an altered version of de Brujin sequences~\cite{Ralston-MM-1982}, as explained in \cite{Benegui-Access-2020}. Rauber et al. \cite{rauber2017foolbox} released a Python-based toolbox used for model robustness benchmarking, called Foolbox. We use the built-in Foolbox attack strategies to generate adversarial examples for our convolutional neural networks.

In summary, our aim is to answer the following research questions (RQs) in the context of adversarial example generation on motion sensor data:
\begin{itemize}
    \item RQ1: Can we generate adversarial examples to attack deep learning models for user identification based on motion sensor data?
    \item RQ2: Are the adversarial examples universal or specific to certain neural architectures?
    \item RQ3: What generative methods produce the most damaging adversarial attacks?
\end{itemize}

We organize the rest of this paper as follows. We discuss related work in Section~\ref{sec_related_work}. We present the machine learning models and adversarial generation techniques in Section~\ref{sec_method}. We describe the experiments with adversarial attacks in Section~\ref{sec_experiments}. Finally, we draw our conclusions in Section~\ref{sec_conclusion}.

%
%
\section{Related Work}
\label{sec_related_work}

There are many works that conducted research on user identification on mobile devices using motion sensor data~\cite{Bo-IPCCC-2014,Buriro-CODASPY-2018,Buriro-ISBA-2017,Canfora-ICETE-2017,Ehatisham-JNCA-2018,Ku-Access-2019,Li-BIBM-2018,Neverova-Access-2016,Shen-Sensors-2016,Shi-WiMob-2011,Sitova-TIFS-2016,Sun-ECML-2017,Vildjiounaite-ICPC-2006,Wang-Access-2019}. Among the broad range of explored methods, some rely on voice and accelerometer-based user recognition \cite{Vildjiounaite-ICPC-2006}, while others perform human movement tracking based on motion sensors~\cite{Sitova-TIFS-2016}. The most recent and best-performing approaches are based on deep learning~\cite{Benegui-Access-2020,Neverova-Access-2016}. So far, researchers explored recurrent neural networks~\cite{Neverova-Access-2016} and convolutional neural networks~\cite{Benegui-Access-2020}. However, none of the previous works investigated the possibility of generating adversarial attacks by adding indistinguishable perturbations to the motion signals recorded on mobile devices.
In this paper, we focus on convolutional neural networks (CNNs), which were shown to outperform Long Short-Term Memory networks in the recent work of Benegui et al.~\cite{Benegui-Access-2020}. The approach proposed in~\cite{Benegui-Access-2020} consists in converting motion sensor signals into images that are provided as input to a convolutional neural network. Although CNN models provide outstanding results on various computer vision tasks~\cite{Georgescu-Access-2019,He-CVPR-2016,Ionescu-CVPR-2016,Simonyan-ICLR-14}, some recent studies~\cite{carlini2017adversarial,chen2017show,narodytska2017simple,papernot2016transferability} showed that adversarial examples can generate sizeable misclassification error rates, indicating that CNNs are not robust to adversarial attacks. For instance, Chen et al.~\cite{chen2017show} showed that adversarial attacks induce an error rate of $95.80\%$ in CNNs trained for image classification. Simple black-box tools used to generate adversarial samples, e.g.~Foolbox~\cite{rauber2017foolbox}, represent practical ways of crafting attacks without requiring any prior knowledge related to the attacked model~\cite{narodytska2017simple}. Papernot et al.~\cite{papernot2016transferability} noted that generator models for adversarial examples can be universal, i.e.~examples optimized on a classification model can easily produce misclassification errors for other classification models with different architectures. We note that the aforementioned studies targeted neural networks for objects and image classification problems. To our knowledge, we are the first to study adversarial attacks on neural networks used in user identification tasks based on motion sensor signals recorded on mobile devices.

%
%
\section{Method}
\label{sec_method}

\subsection{Convolutional Neural Networks}

With the aim of answering our research questions, we implement four CNN models, with the same design and configuration proposed by Benegui et al.~\cite{Benegui-Access-2020}, starting from a shallow 4-layer CNN architecture and increasing the depth gradually to 6 layers, 9 layers and 12 layers, respectively. Each model consists of different convolutional (conv) layers followed by max pooling and fully-connected (fc) layers. All layers are activated by Rectified Linear Units (ReLU)~\cite{Nair-ICML-2010}, except for the classification layer. Each fc layer has a dropout rate of $0.4$~\cite{Srivastava-JMLR-2014} to prevent overfitting. Max pooling layers have a pool size of $2\times2$ and are applied at stride $2$. Since our models are trained for multi-class classification, Softmax activation is preferred in the last layer, since it outputs the probability of each class. Network training is performed using the Adam optimizer~\cite{Kingma-ICLR-2015} with the categorical cross-entropy loss function. In all models, the final layer is composed of 50 neurons, equal to the number of classes.

The 4-layer architecture is composed of 1 conv layer with 32 filters applied at a stride of $1$, followed by 2 fc layers and the softmax activation layer. In the 6-layer CNN, we employ 3 conv layers, 2 fc layers and one Softmax classification layer. The third architecture, with a depth of 9 layers, is built as follows: 6 conv layers, 2 fc layers and the classification layer. Lastly, our 12-layer CNN is composed of 9 conv layers, 2 fc layers and the Softmax classification layer. While the four models share some architectural elements, each model is trained independently, having no weights in common with any of the other models.

\subsection{Adversarial Example Generation Strategies}

Adversarial examples represent scarcely modified data samples that, when provided as input to neural networks, determine incorrect predictions.
Starting from the assumption that different methods of generating adversarial examples can provide different results, we propose to experiment with four different gradient-based or decision-based attack methods. The goal of crafting an adversarial example from a real example is to produce a misclassification error, i.e.~the adversarial attack is about inducing incorrect predictions. This is achieved by maximizing the categorical cross-entropy loss. In order to craft a realistic adversarial example, we also aim at minimizing the mean squared error with respect to the original example used as starting point for the optimization. Therefore, our second goal is to create adversarial samples as similar as possible to the originals. By jointly maximizing the categorical cross-entropy and minimizing the mean squared error between two inputs (the original and the adversarial example), we can obtain an adversarial example with a high percentage of similarity to the original, unnoticeable for humans, that is wrongly classified by the neural network. We note that the generator might not always converge to an adversarial example having both properties, i.e.~not all attacks are successful. Next, we briefly present the considered attacks.

\begin{figure}[!t]
\centering
\includegraphics[width=1.0\textwidth]{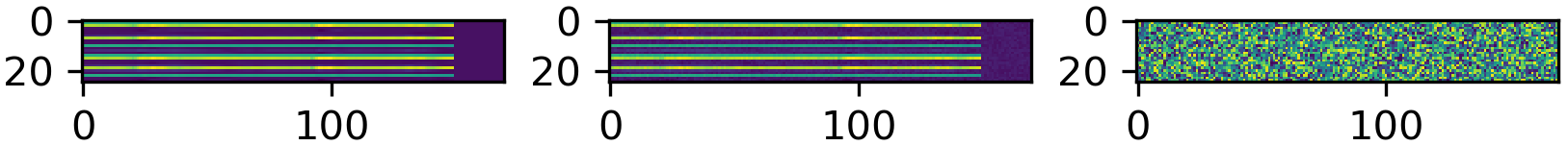}
\caption{An original and an adversarial example generated by the Boundary Attack. The left-most image represents the original motion sensor signals. The second image is the adversarial example. The right-most image depicts an exaggerated map of differences between the original and the adversarial image. Best viewed in color.} \label{fig_adversarial_sample}
\end{figure}

\noindent
{\bf Boundary Attack.}
The Boundary Attack \cite{brendel2017decision} method is a simple and efficient attack that starts from a randomly generated perturbation vector. In each iteration, the method regenerates the adversarial sample until the difference to the original input becomes smaller, yet still recognized as adversarial, by minimizing the $L_2$-norm of perturbations. Figure~\ref{fig_adversarial_sample} shows a comparison between an original example and an adversarial example generated by the Boundary Attack.


\noindent
{\bf Fast Gradient Sign Attack.}
This attack perturbes the input sample given to a classification model considering the gradient of the loss, gradually amplifying the gradient magnitude until the model misclassifies the generated example \cite{goodfellow2014explaining}. 



\noindent
{\bf DeepFool $\mathbf{L_2}$ Attack.}
Introduced by Moosavi-Dezfooli et al. \cite{moosavi2016deepfool}, the DeepFool $L_2$ attack takes an original sample and generates a perturbation vector that is added to the adversarial sample. For each class, the algorithm computes the smallest distance required to reach the class boundary using a linear classifier. Once the smallest distance has been computed, a corresponding move towards the direction of the class boundary is made. Using this attack, a perturbation vector that misleads the classifier can be identified in as less as three iterations. 


\noindent
{\bf Jacobian-Based Saliency Map Attack.}
Papernot et al. \cite{papernot2016cleverhans} introduced a targeted attack based on gradients, that computes a score, called the saliency score, for each feature available in the original sample. The computations are performed in an iterative manner. Therefore, in each iteration the saliency score can change the prediction of the model from the original class to a different class. 




%
%
\section{Experiments}
\label{sec_experiments}

\subsection{Data Set}

We use the HMOG data set \cite{Sitova-TIFS-2016}, which is composed of discrete signals, gathered from the accelerometer and gyroscope motion sensors, while 50 users perform tap gestures on a mobile device screen. While each user is performing a tap gesture, signals are recorded for a period of $1.5$ seconds, as follows: values are captured starting with $0.5$ seconds before the tap, continuing for another second after the tap was performed. Sensor sample values are arranged on 3 spatial coordinates, namely $(x,y,z)$. Since the values are recorded at an approximate frequency of 100 Hz, we commonly obtain around 150 values for the signal corresponding to one axis. For our experiments, we recorded signals associated with 200 gestures per user, thus obtaining a data set of 10.000 samples (50 users $\times$ 200 samples).
For the training and evaluation process, we employ an $80\%$--$20\%$ train-test split, hence, 160 samples per user are used for training, while the other 40 samples are used for testing. Adopting a multi-class user identification model on the set of users, we obtain 8.000 training samples and 2.000 test samples. The test samples are used to generate adversarial examples. After the generative method reaches a convergence point for each test sample, the generated adversarial examples are fed into the neural models to reassess the performance of the user identification models under adversarial attacks.

\subsection{Evaluation metrics}

In our experiments, we employ the misclassification error 
as evaluation metric. Since adversarial samples are crafted to fool the deep CNNs, in an attempt to make the models assign labels different from the expected ones, the goal of the attack is to increase the misclassification rate. 
We also calculate the misclassification error of the deep networks on the original test samples, thus providing a measure of the impact of the adversarial attacks. 


\subsection{Parameter Tuning and Implementation Details}\label{subsec_implementation}

Benegui et al.~\cite{Benegui-Access-2020} described the optimal hyperparameters for the CNN models trained on the HMOG data set. Hence, we employ the same hyperparameters, obtaining similar results. The learning rate is $10^{-3}$ and the mini-batch size is $32$. Furthermore, we use dropout layers with a rate of $0.4$ after each fully-connected layer. We train each CNN model using the Adam optimizer \cite{Kingma-ICLR-2015} for a total of 50 epochs. We use the Foolbox library~\cite{rauber2017foolbox} to generate adversarial samples. Foolbox provides out-of-the-box implementations for state-of-the-art gradient-based and decision-based attacks. We employ the four example generation methods presented in Section \ref{sec_method} using the default Foolbox hyperparameters, except for the Boundary Attack method, where we employ a batch size equal to $20$ and run the algorithm for a maximum of $5$ iterations on each original sample.

\subsection{Experimental Setup}

After neural network training on 160 samples per user, the remaining 40 data samples are used for both testing and adversarial generation. Since our data set contains 50 users in total, we generate 2.000 adversarial samples (40 test samples $\times$ 50 users) with each generative method, namely the Fast Gradient Sign Attack, the DeepFool $L_2$ Attack, the Jacobian-Based Saliency Map Attack and the Boundary Attack. Thus, we obtain a total of 8.000 adversarial examples. In our experiments, we employ four CNN architectures of different depths, particularly of 4 layers, 6 layers, 9 layers and 12 layers. To answer RQ2, we test the adversarial examples generated for one CNN model on the other CNN models.


Lastly, we explore an ensemble classification method, by combining all the proposed CNNs through a weight voting scheme, using the classification accuracy rates of the independent models as weights. In this context, an attack is universal if the majority vote predicted by the ensemble is not equal to the ground-truth label. Adversarial samples are given as input to the ensemble, with the aim of further exploring their universality. 

\subsection{Results}


\begin{table}[!t]
\caption{Empirical results obtained by distinct neural network architectures under various adversarial attacks, in the user identification task. The results are reported in terms of the misclassification error (lower values indicate higher robustness to adversarial attacks). The model for which the adversarial examples were generated is marked with an asterisk.}\label{tab_results}
\centering
\begin{tabular}{|l|c|c|c|c|}
\hline
                                        & \multicolumn{4}{|c|}{Misclassification Error} \\ 
\cline{2-5} 
Adversarial Generation Strategy         & 4-layer               & 6-layer           & 9-layer               & 12-layer \\ 
                                        & CNN                   & CNN               & CNN                   & CNN \\ 

\hline
\hline
None                                    & $8.25\%$            & $8.20\%$            & $4.45\%$            & $5.80\%$ \\
\hline
Boundary Attack on 4-layer CNN          & $100.00\%^\star$    & $39.50\%$           & $34.60\%$           & $25.90\%$ \\
DeepFool $L_2$ Attack on 4-layer CNN    & $100.00\%^\star$    & $6.45\%$            & $6.00\%$            & $6.50\%$ \\
Gradient Sign Attack on 4-layer CNN     & $99.50\%^\star$     & $14.40\%$           & $11.55\%$           & $13.05\%$ \\ 
Saliency Map Attack  on 4-layer CNN     & $100.00\%^\star$    & $82.85\%$           & $60.80\%$           & $30.80\%$ \\
\hline
Boundary Attack on 6-layer CNN          & $79.40\%$           & $100.00\%^\star$    & $48.60\%$           & $42.30\%$ \\
DeepFool $L_2$ Attack on 6-layer CNN    & $9.30\%$            & $100.00\%^\star$    & $6.20\%$            & $8.00\%$ \\
Gradient Sign Attack on 6-layer CNN     & $22.65\%$           & $99.95\%^\star$     & $16.90\%$           & $25.20\%$ \\
Saliency Map Attack on 6-layer CNN      & $91.95\%$           & $100.00\%^\star$    & $66.55\%$           & $43.30\%$ \\
\hline
Boundary Attack on 9-layer CNN          & $81.90\%$           & $67.10\%$           & $99.95\%^\star$     & $47.40\%$ \\
DeepFool $L_2$ Attack on 9-layer CNN    & $8.45\%$            & $5.25\%$            & $100.00\%^\star$    & $8.85\%$ \\
Gradient Sign Attack on 9-layer CNN     & $17.85\%$           & $17.90\%$           & $100.00\%^\star$    & $22.05\%$ \\
Saliency Map Attack on 9-layer CNN      & $89.25\%$           & $85.55\%$           & $100.00\%^\star$    & $53.95\%$ \\
\hline
Boundary Attack on 12-layer CNN         & $83.40\%$           & $72.05\%$           & $64.55\%$           & $100.00\%^\star$ \\
DeepFool $L_2$ Attack on 12-layer CNN   & $8.45\%$            & $5.30\%$            & $6.65\%$            & $100.00\%^\star$ \\
Gradient Sign Attack on 12-layer CNN    & $16.85\%$           & $18.20\%$           & $11.55\%$           & $99.8\%^\star$ \\
Saliency Map Attack on 12-layer CNN     & $93.20\%$           & $89.35\%$           & $83.60\%$           & $100.00\%^\star$ \\
\hline
\end{tabular}
\end{table}

In Table~\ref{tab_results}, we present the empirical results obtained by distinct neural network architectures in the user identification task, under various adversarial attacks. One by one, each CNN model is selected as the attacked model. Since we are interested in the universality of the adversarial examples, we feed the adversarial examples generated for a specific model to all other models. To allow a better estimation of the impact of each adversarial attack, we report the misclassification error rates on the original test samples.

\noindent
{\bf Adversarial examples optimized for the 4-layer CNN.}
We observe that, when the attacks are designed for the 4-layer CNN model, the misclassification rate for three of the attacks (DeepFool $L_2$, Jacobian-Based Saliency Map and Boundary) is $100\%$. The Fast Gradient Sign Attack induces a misclassification rate of $99.50\%$. We thus conclude that the 4-layer CNN is not robust to adversarial attacks optimized on itself. However, when we feed the adversarial examples to deeper architectures, depending on the attack strategy, we observe that the misclassification error varies considerably, attaining values between a minimum of $6.00\%$ and a maximum of $82.85\%$. In the cross-model evaluation, we note that the DeepFool $L_2$ Attack is not universal, the deeper CNNs being robust to the examples generated by DeepFool $L_2$. In terms of universality, the most successful seems to be the Jacobian-Based Saliency Map Attack, although its effectiveness degrades as the CNN model under attack gets deeper. For example, the Jacobian-Based Saliency Map Attack increases the missclasification error of the 12-layer CNN from $5.80\%$ to $30.80\%$, most of the attacks being rejected by the CNN model.

\noindent
{\bf Adversarial examples optimized for the 6-layer CNN.}
When the attacks are specifically optimized for the 6-layer CNN, its misclassification error rates are $100\%$ or nearly $100\%$. Just as the 4-layer CNN, the 6-layer CNN is not robust to adversarial examples optimized on itself. Except for the DeepFool $L_2$ Attack, we observe that the adversarial examples increase the misclassification error of the remaining architectures. The least affected model by the DeepFool $L_2$ Attack is the 9-layer CNN, attaining a misclassification error of $6.20\%$. For the Boundary Attack, it seems that the impact of the adversarial examples is inversely proportional to the depth of the neural networks. The most effective attack across all models seems to be the Jacobian-Based Saliency Map Attack. For example, the 4-layers CNN gives a misclassification error of $91.95\%$ when it is attacked with the adversarial examples generated by the Jacobian-Based Saliency Map method applied on the 6-layer CNN. An interesting observation is that the adversarial examples optimized for the 6-layer CNN seem to have a higher degree of universality than the adversarial examples optimized for the 4-layer CNN.

\noindent
{\bf Adversarial examples optimized for the 9-layer CNN.}
The adversarial examples generated for the 9-layer CNN are very effective on the 9-layer CNN, just as we observed in the previous cases. Indeed, the misclassification error rates are typically close to $100\%$, indicating that the 9-layer CNN is not robust to adversarial examples. As we observed in the previous experiments, many of the adversarial attacks designed for the 9-layer CNN raise the misclassification error rates for the remaining models by considerable margins. The most affected models seem to be the 4-layer CNN and the 6-layer CNN. Once again, the least effective attack is the DeepFool $L_2$ Attack, while the most effective one is the Jacobian-Based Saliency Map Attack.


\begin{table}[!t]
\caption{Empirical results obtained by the CNN ensemble based on weighted majority voting under various adversarial attacks, in the user identification task. The results are reported in terms of the misclassification error (lower values indicate higher robustness to adversarial attacks).}\label{tab_results_ensemble}
\centering
\begin{tabular}{|l|c|}
\hline
Adversarial Generation Strategy     & Misclassification Error of CNN Ensemble \\
\hline
\hline
None                                    & $2.45\%$ \\
\hline
Boundary Attack on 4-layer CNN          & $24.50\%$ \\
DeepFool $L_2$ Attack on 4-layer CNN    & $2.79\%$ \\
Gradient Sign Attack on 4-layer CNN     & $8.50\%$ \\ 
Saliency Map Attack  on 4-layer CNN     & $48.10\%$ \\
\hline
Boundary Attack on 6-layer CNN          & $46.09\%$ \\
DeepFool $L_2$ Attack on 6-layer CNN    & $2.95\%$ \\
Gradient Sign Attack on 6-layer CNN     & $15.04\%$ \\
Saliency Map Attack on 6-layer CNN      & $59.15\%$ \\
\hline
Boundary Attack on 9-layer CNN          & $59.25\%$ \\
DeepFool $L_2$ Attack on 9-layer CNN    & $3.10\%$ \\
Gradient Sign Attack on 9-layer CNN     & $15.84\%$ \\
Saliency Map Attack on 9-layer CNN      & $77.40\%$ \\
\hline
Boundary Attack on 12-layer CNN         & $65.94\%$ \\
DeepFool $L_2$ Attack on 12-layer CNN   & $3.04\%$ \\
Gradient Sign Attack on 12-layer CNN    & $11.05\%$ \\
Saliency Map Attack on 12-layer CNN     & $86.75\%$ \\
\hline
\end{tabular}
\end{table}

\noindent
{\bf Adversarial examples optimized for the 12-layer CNN.}
So far, the most robust CNN model, with respect to all the considered attacks, was the 12-layer CNN. However, when the adversarial examples are specifically crafted for the 12-layer CNN, the model is no longer robust, attaining misclassification rates of $100\%$ or close. Furthermore, the impact of the adversarial examples generated with the Jacobian-Based Saliency Map method on the shallower CNN models is major. Indeed, the Jacobian-Based Saliency Map Attack induces misclassification error rates between $83.60\%$ for the 9-layer CNN and $93.20\%$ for the 4-layer CNN. On the other hand, the DeepFool $L_2$ Attack is not effective.

\noindent
{\bf Generic observations.}
Considering the overall results, we observe that the adversarial examples generated using the Jacobian-Based Saliency Map are the most effective across CNN models. In general, the attacks seem to have a higher degree of universality as the adversarial examples are optimized for deeper networks. In the same time, it seems that the depth of the network under attack is strongly correlated to the robustness to adversarial examples optimized on other CNN models, i.e.~deeper models are more robust. However, each CNN model can be easily attacked when the adversarial examples are optimized on the respective model, proving that no model is robust to targeted attacks. In this context, we next seek to find out if an ensemble formed of the four CNN models is robust to all adversarial attacks.

\noindent
{\bf Adversarial attacks on CNN ensemble.}
In our last experiment, we feed the adversarial examples to an ensemble of the four CNN models, presenting the corresponding results in Table~\ref{tab_results_ensemble}. First, we note the ensemble attains improved results on original test samples, the error rate being $2\%$ lower than that of the best individual model, namely the 9-layer CNN. While DeepFool $L_2$ and Fast Gradient Sign do not seem to break the ensemble, it appears that the CNN ensemble is not robust to the Boundary Attack and the Jacobian-Based Saliency Map Attack. The misclassification error rates are typically higher as the adversarial attacks are targeted on deeper and deeper networks. This is consistent with the results presented in Table~\ref{tab_results}. We conclude that combining the CNNs into an ensemble is not effective for blocking certain adversarial attacks.


%
%
\section{Conclusion}
\label{sec_conclusion}

In this paper, we studied different methods of generating adversarial examples for deep models trained on motion sensor signals. We conducted a series of user identification experiments, considering different CNN architectures  using both original and adversarial examples. The evaluation led to conclusive empirical results, enabling us to answer the proposed research questions. We hereby conclude our work by answering the research questions:
\begin{itemize}
\item RQ1: Can we generate adversarial examples to attack deep learning models for user identification based on motion sensor data? \\
Answer: Generating adversarial attacks is possible, producing misclassification error rates of nearly $100\%$, irrespective of the CNN architecture. The answer to RQ1 is affirmative.
\item RQ2: Are the adversarial examples universal or specific to certain neural architectures? \\
Answer: After experimenting with four adversarial attack strategies, we observed that some attacks are specific, namely the DeepFool $L_2$ Attack and the Fast Gradient Sign Attack, and others are rather universal, namely the Boundary Attack and the Jacobian-Based Saliency Map Attack. In conclusion, it really depends on the attack strategy. It is perhaps important to mention here that the degree of universality grows with the depth of the target CNN used to optimize the adversarial examples.
\item RQ3: What generative methods produce the most damaging adversarial attacks?
\\Answer: All the consider adversarial attack methods induce misclassification error rates of nearly $100\%$ when the optimization and the evaluation is performed on the same CNN model. However, considering the universality of the attack strategies, the most damaging adversarial attack is definitely the one based on the Jacobian-Based Saliency Map.
\end{itemize}

In summary, we conclude that deep neural networks for user identification are affected by adversarial examples, even in the cross-model evaluation setting. In most cases, the misclassification error rates grow by substantial margins. In future work, we aim to design novel deep neural networks that are robust to adversarial examples, perhaps by introducing a way to identify adversarial examples.

%
%

\bibliography{references}{} 
\bibliographystyle{splncs04}

\end{document}